\documentclass[letterpaper, 10 pt, conference]{ieeeconf}  

\IEEEoverridecommandlockouts                              

\usepackage{graphicx} 
\usepackage{amsmath}
\usepackage{array} 
\newcolumntype{C}[1]{>{\centering\arraybackslash}p{#1}} 

\title{\LARGE \bf
Dynamic Agentic AI Expert Profiler System Architecture for Multidomain Intelligence Modeling
}

\author{
Aisvarya Adeseye$^{1}$, 
Jouni Isoaho$^{1}$, 
Seppo Virtanen$^{1}$, 
Mohammad Tahir$^{1}$%
\thanks{*This work was not supported by any organization}%
\thanks{$^{1}$Department of Computing, University of Turku, Turku, Finland
{\tt\small \{aisvarya.a.adeseye, jouni.isoaho, seppo.virtanen, tahir.mohammad\}@utu.fi}}%
}

\begin{document}

\maketitle
\thispagestyle{empty}
\pagestyle{empty}

\begin{abstract}
In today's artificial intelligence (AI) driven world, modern systems communicate with people with diverse backgrounds and skillsets. Hence, for this human-machine interaction to be more meaningful, it is important for machines to be aware of the context and user expertise. Therefore, this study proposes an agentic AI profiler that uses four levels: Novice, Basic, Advanced, and Expert to classify natural language responses. The system was built with a modular layered architecture and LLaMA v3.1 (8B) with components for text preprocessing, scoring, aggregation, and classification. The system evaluation was done in 2 phases: static and dynamic. 
The static phase analyzed pre-recorded transcripts from 82 participants. Similarly, the dynamic phase evaluated 402 live interviews conducted with an agentic AI interviewer. In both phases, participants’ self-ratings were compared with the profiler's rating. However, during the dynamic phase, participants' expertise was evaluated after every response, not at the end of the interview as in the static phase. Across domains, 83\%–97\% of profiler evaluations matched participants’ self-evaluations. The remaining differences came from participants overrating or underrating themselves, unclear responses, and occasional LLM misinterpretations of nuanced expertise.
\end{abstract}

\section{Introduction}

Modern AI systems, especially Large Language Models (LLMs), can understand simple human language. They can also derive meanings from human responses and provide answers that fit the situation \cite{b1, adeseye2025llmqual}. Because of this, machines can now communicate in a way that feels more natural and human. This helps people learn, share ideas, and work better in many areas, such as education, healthcare, and business \cite{b2}.
The ability of LLMs to adapt to a particular language, tone or intent is a major strength \cite{b3, adeseye2025promptframework}. However, for effective communication, AI systems must understand the user's expertise level to adapt communication style \cite{b4}, so that it provides meaningful support that builds user trust \cite{b5}. 

Therefore, this study aims to develop and validate a profiling system that is capable of dynamically and transparently analyzing human expertise with a local LLM. When natural language capabilities of LLMs are combined with expertise level profiling, it enables LLM-driven systems to exchange thoughts at a more personalized level \cite{b6,b7,b8,b9}.



Thus, this research is guided by the key question:
\textit{"How can we design an AI-based system architecture that accurately evaluates and classifies human expertise from different domains?"}



The main contribution of the study is the design and validation of an expertise profiler system architecture. This is a framework that integrates linguistic scoring, reasoning analysis, and contextual evaluation in one unified process. It introduces a new approach to test expertise in two modes: static and offline mode, and dynamic and real-time mode. Hence, adaptive profiling helps reduce bias and improve accuracy caused by the way people perceive themselves. Additionally, it provides an explainable and interpretable structure, enhancing the transparency and fairness of AI-driven evaluation systems. 




\section{Related Works}

Borna et al. \cite{b10} built a Machine Learning (ML) model that processes medical data to find experts. It has layers for data cleaning, extracting features, and classifying experts. It was mainly used to match doctors with tasks. The raising of privacy issues because of patient data is a limitation. The work of Bukowski et al. \cite{b11} focused on using text mining to build profiles of emerging experts. The system scans publications and patients to find keywords and topics. It comprises data collection, text analysis, and a profile creation module. Limitations include not being able to capture experts with hidden skills and little publication.
Also, Balog and Rijke \cite{b12} built a probabilistic model with similarity measures to match expert profiles inside organizations. The system architecture includes indexing, retrieval, and a ranking component. Drawbacks included the need for a large data set and structured company data.

Similarly, Fazel-Zarand and Fox \cite{b13} built a time-aware recommender system and temporary models to track skillset change over a period of time for staff recruitment and career planning in companies. When skillsets change quickly, it fails to cope, which is a major limitation.
Finally, Fu and Akbar \cite{b14} used a co-authorship graph network to group and cluster experts. It is used for research community mapping and leadership identification. Some limitation includes overdependence on complete data and difficulty in mapping overlapping communities. 
The literature indicates that there are no expert profilers that analyze human textual conversations to determine expertise level, which is the focus of this study.


\section{System Architecture of Expert Profiler}

\begin{figure*}[!t]
\centering
\includegraphics[width=1\textwidth]{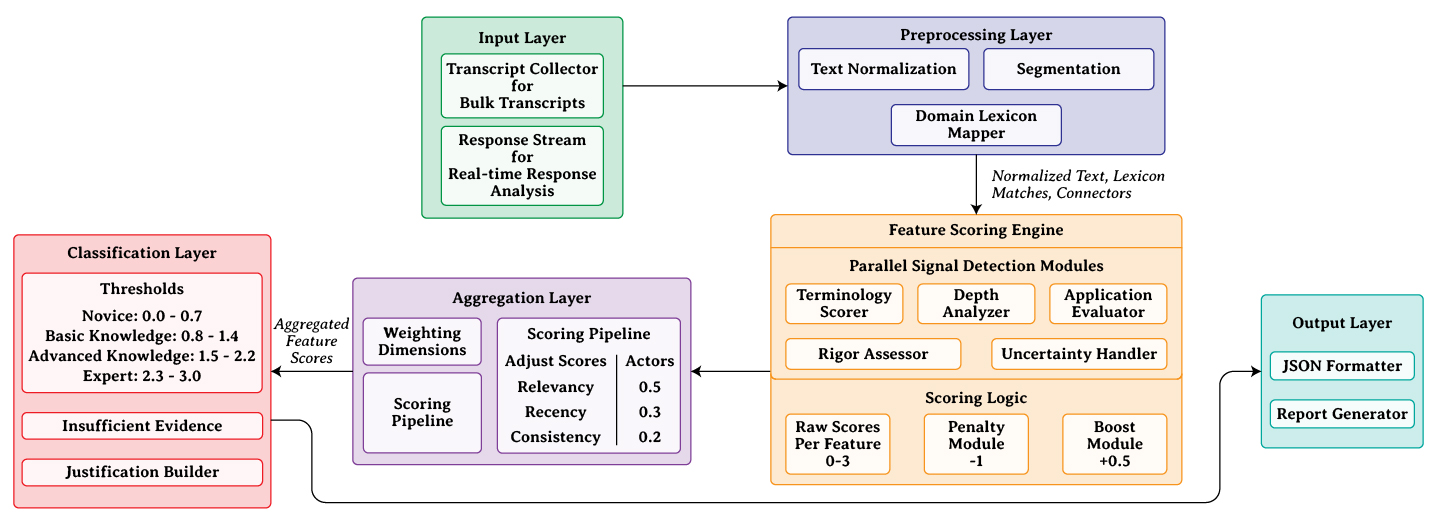}
\vspace{-0.25cm}
\caption{Expert Profiler System Architecture: A Layered Framework for Expertise Classification from Textual Responses}
\vspace{-0.5cm}
\label{fig:ExpertProfilerArchitecture}
\end{figure*}

The Expert Profiler is an agentic AI system that uses written textual conversations like interview transcripts or live discussions to determine the expertise level of a person. The system is divided into several sequential but modular layers with their own responsibilities. Thus, it is scalable, transparent, and flexible, making it adaptable in various areas such as education, health, and research. The following section explains the architecture in Figure \ref{fig:ExpertProfilerArchitecture}.

\subsection{Input Layer}
This is the first layer; it takes inputs in two ways, thus making it versatile:
\begin{itemize}
    \item \textbf{Transcript Collector:} It collects pre-recorded large interview transcripts for batch processing.
    \item \textbf{Response Stream:} It collects live responses for real-time analysis during interview discussions.
\end{itemize}


\subsection{Preprocessing Layer}
Raw conversational data contains noises that include inconsistencies and irrelevant patterns. Consequently, this layer eliminates such noises using three key processes:
\begin{enumerate}
    \item \textbf{Text Normalization:} This focuses on eliminating formatting differences. For example, treating “AI,” “A.I.,” and “artificial intelligence” as the same thing. This eliminates confusion and improves consistency.

    \item \textbf{Segmentation:} This process involves breaking long transcripts into shorter sentences and phrases to analyze them as separate chunks and not the entire transcript at once. This improves textual understanding.
     
    \item \textbf{Domain Lexicon Mapper:} This includes defining domain-specific vocabulary that helps in improving the system's transcript understanding. For example, words with more than one meaning, such as "token" and "fine-tuning" can be explicitly stated to the system as technical AI terms.


    .

\end{enumerate}
This ensures that the analysis is unbiased as a result of linguistic irregularities, enhancing interpretability, which is important for experts profiling in different domains.


\subsection{Feature Scoring Engine}
This is the core of the system. It contains five parallel modules that cover different expertise indicators.


\begin{itemize}
    \item \textbf{Terminology Scorer:} This is a measure of how domain-specific vocabulary is applied accurately by the participants. For example, in cybersecurity, “phishing”, “encryption”, and “firewall” are accurate domain terminology.  
    


    \item \textbf{Depth Analyzer:} This measures how well the ideas are linked together and not just surface definitions, for example, an explanation of \textbf{how} encryption works and \textbf{why} it is important, shows deeper understanding.
    

    \item \textbf{Application Evaluator:} This is a measure of the application of knowledge to real-life situations. For example, explaining \textbf{how employee data could be protected} indicates applied understanding.
    

    \item \textbf{Rigor Assessor:} Involves the application of logical reasoning to check the structure, clarity, and evidence presented in the responses.

    \item \textbf{Uncertainty Handler:} This focuses on finding phrases that indicates uncertainty like “\textit{I think}” or “\textit{maybe}” as well as strong claims that lacks proofs.
    
\end{itemize}

Each feature is scored between \textbf{0} and \textbf{3} by the LLM.  The \textbf{average feature score} is then computed as follows:

{\scriptsize
\[
\textbf{Score}_{\text{avg}} = \frac{\text{Terminology} + \text{Depth} + \text{Application} + \text{Rigor} + \text{Uncertainty}}{5}
\]
}
Then, two adjustments are applied to it: a \textbf{Penalty (-1)} or a \textbf{Boost (+0.5)}.

\begin{itemize}
    \item \textbf{Penalty:} If an incorrect response was provided, for example, saying “An LLM is a type of electric vehicle”, then 1 point is deducted.
    
    {\scriptsize
    \[
    \textbf{Score}_{\text{penalised}} = \max(0,\; \textbf{Score}_{\text{avg}} - 1)
    \]
    }
    
    \item \textbf{Boost:} If a factually correct response is provided, for example, saying "personal details from interview transcript should be anonymized", then  \textbf{0.5} point is added.
    
    {\scriptsize
    \[
    \textbf{Score}_{\text{boosted}} = \min(3,\; \textbf{Score}_{\text{avg}} + 0.5)
    \]
    }
\end{itemize}

These adjustments ensure that the scoring remains fair, realistic, and aligned with the quality of the response. The average feature score is computed to create a single quality measure for each response by combining all five feature indicators. The Penalty and Boost adjustments are then applied to this average Score\textsubscript{avg}. The adjusted average score is used to flag responses as either unreliable when very low or strongly valid when high. This ensures that the corrections are applied to the entire response equally. After this adjustment step, the individual feature scores (Terminology, Depth, Application, Rigor, and Uncertainty) are passed to the Aggregation Layer (Relevancy, Recency, and Consistency) to compute the dimension scores. The average score itself is not directly used in the final expertise calculation.

\subsection{Aggregation Layer}
This layer combines individual feature scores using weighted dimensions. By Weighted dimensions, we mean that each factor does not contribute equally; instead, each one has a multiplier that reflects its importance when computing the final score. 



\begin{itemize}
    \item \textbf{Relevancy (weight = 0.5):} This measures how relevant and meaningful the response is to the questions asked, reflecting how well the participant stays on topic and applies knowledge. It is computed from the average of \textit{Terminology} and \textit{Application} feature scores across responses:

    {\scriptsize
    \[
    Relevancy = \text{avg}(Terminology, Application)
    \]
    }

    \item \textbf{Recency (weight = 0.3):} Measures how up-to-date the knowledge is. Recent terms and concepts receive higher scores. It is computed from the average of \textit{Terminology} and \textit{Depth} feature scores:

    {\scriptsize
    \[
    Recency = \text{avg}(Terminology, Depth)
    \]
    }
    
    \item \textbf{Consistency (weight = 0.2):} Measures how stable and coherent answers are across questions. Experts use terms and logic consistently. For example, confusing prompt engineering with fine-tuning lowers this score. It is computed from the average of \textit{Rigor} and \textit{Uncertainty} feature scores:

    {\scriptsize
    \[
    Consistency = \text{avg}(Rigor, Uncertainty)
    \]
    }
    
\end{itemize}
Each dimension is also scored between \textbf{0} and \textbf{3}. The \textbf{final expertise score} is a weighted average:

\vspace{-0.2CM}
{\scriptsize
\[
\textbf{Score}_{\text{final}} = 0.5 \times \textbf{Relevancy} + 0.3 \times \textbf{Recency} + 0.2 \times \textbf{Consistency}
\]
}
Because the sum of the weights equals 1 and each dimension is between 0 and 3,  Score\textsubscript{final} always stays in the 0-3 range. The weights were empirically chosen by identifying which dimensions were more important to the expert profiler; the chosen values reflect the importance of relevance, timeliness, and coherence. They help reflect not only what the user knows but also how current, relevant, and stable that knowledge is.



\subsection{Classification Layer}
This is the decision-making part of the expert profiler. All feature scores are combined into a final score that reflects the overall participant's expertise level. The participant is then placed into four possible groups: Novice, Basic Knowledge, Advanced Knowledge, or Expert. Each range depicts a different level of reasoning ability and knowledge that reflects what the participant knows and how well they use what they know.

\begin{itemize}

    \item \textbf{Novice (0.0–0.7):} This category of users is characterized by the following: providing short and unclear responses, very limited knowledge, weak reasoning despite the use of basic terms, and a lack of practical understanding. The lower limit helps the system identify complete beginners and guessers.
    
   .

    \item \textbf{Basic Knowledge (0.8–1.4):} 
    This includes participants who exhibit basic awareness of the topic; they can provide simple definitions but cannot connect them to real-life situations. This represents people who are not beginners but are starting to build understanding.
    

    \item \textbf{Advanced Knowledge (1.5–2.2):} 
    This includes participants exhibiting strong and clear reasoning and understanding by analyzing, comparing, and applying ideas with few mistakes. The upper bound contains participants close to becoming an expert.
    .

    \item \textbf{Expert (2.3–3.0):} This category is for participants demonstrating deep understanding, confidence, and accuracy. Their responses are clear, well-structured, and relevant. They are also supported with examples or evidence. The 2.3 threshold ensures that only consistently high performers attain this level. 

\end{itemize}
This threshold-based classification ensures clarity and transparency in the evaluation process. Also, this layer has other essential parts as described below

\begin{itemize}


    \item \textbf{Insufficient Evidence Detection:}This protects the system's credibility, identifying when the data or evidence is not enough to make a valid and reliable classification. For example, when a transcript has only a few responses.

    \item \textbf{Justification Builder:} This explains the rationale that governs the final expertise level classification from the features, weighting dimensions, and scoring pipeline. For example, a justification of "Advanced knowledge" could be written as: the participant correctly applied domain-specific terminology but had small reasoning gaps.

\end{itemize}

This layer improves the reliability and accountability of the system, which is necessary if fairness and explainability are to be ensured, which are critical parts of modern AI-based assessment systems.


\subsection{Output Layer}
This layer is responsible for providing outputs that are useful to humans and machines. It includes:

\begin{itemize}
    \item \textbf{JSON Formatter:} Formats the results, making it structured, machine-readable, and integration-ready. The output is created from feature scores, final categorization, confidence level, and justification.


    \item \textbf{Report Generator:} 
    This creates a friendly, but detailed summary explaining the rationale behind the classification result, so that non-technical users can understand it. For example, part of the report might read: "The participant exhibited strong and consistent understanding and reasoning with minor mistakes. Advanced Knowledge is an appropriate classification".


\end{itemize}
This structure enables automation and interpretation of results, making it meaningful to both machines and humans.




\section{Application and Validation}
The expert profiler was applied to two different research topics and settings (real-time and offline interviews). This helps in testing the effectiveness of the profiler to evaluate human expertise across different topics; it compares its rating to how the participants rated themselves. The system architecture was implemented with a LLaMA v3.1 (8B) analytical pipeline covering pre-processing, feature scoring, aggregation, and classification. LLaMA v3.1 (8B) was selected because our previous study \cite{adeseye2025efficientprompt} showed that it provides strong and reliable results compared to other local models. It achieves good performance without requiring large computing resources. This makes it suitable for resource-constrained environments, where memory and processing power are limited.

\subsection{Static Expert Profiling and Offline Validation}
The first research topic was an already completed face-to-face interview with 82 participants about the security and privacy concerns with the introduction of gamification in an organizational setting. The expert profiler analyzed pre-collected transcript. First, the participants rated their expertise level prior to the beginning of the interview. Secondly, the expert profiler processes each transcript to determine the participant's expertise.




A comparison of these evaluation methods is presented in Table~\ref{tab:profiler_case_study}.

From table \ref{tab:profiler_case_study}, it is clear that the profiler expertise rating usually matches how the participants rated themselves. This is particularly obvious for the security case study, where it achieved 98\%, indicating that it works well for technical topics. However, for privacy and gamification, the differences varied further. This indicates that for broader topics that depend more on interpretation and personal experience, the profiler found them harder.

\vspace{-0.2CM}
\begin{table}[!htbp]
\centering
\caption{Comparison of Expertise Levels Between Self-Evaluation and Profiler final report in \%}
\label{tab:profiler_case_study}
\scriptsize
\setlength{\tabcolsep}{2pt}
\renewcommand{\arraystretch}{1.1}

\begin{tabular}{|p{0.19\linewidth}|C{0.085\linewidth}|C{0.085\linewidth}|C{0.085\linewidth}|C{0.085\linewidth}|C{0.085\linewidth}|C{0.085\linewidth}|C{0.085\linewidth}|}
\hline
\multicolumn{1}{|c|}{\textbf{Case Study}} &
\multicolumn{1}{c|}{\textbf{Same}} &
\multicolumn{3}{c|}{\textbf{Profiler Higher}} &
\multicolumn{3}{c|}{\textbf{Profiler Lower}} \\
\cline{3-8}
\multicolumn{1}{|c|}{} &
\multicolumn{1}{c|}{} &
\textbf{H1} & \textbf{H2} & \textbf{H3} &
\textbf{L1} & \textbf{L2} & \textbf{L3} \\ \hline

Security     & 98 & 2 & 0 & 0 & 0 & 0 & 0 \\ \hline
Privacy      & 89 & 2 & 1 & 0 & 5 & 3 & 0 \\ \hline
Gamification & 83 & 8 & 2 & 1 & 5 & 1 & 0 \\ \hline
\end{tabular}

\vspace{0.2cm}
{\textit{Note: H1, H2, and H3 indicate that the Profiler rated participants 1, 2, or 3 levels higher than their self-evaluation. L1, L2, L3 indicate the Profiler rated them 1, 2, or 3 levels lower.}}
\vspace{-0.4CM}
\end{table}

\subsection{Dynamic Expert Profiling, Real-Time Interview Adaptation, and Validation }

The second research setting was an agentic AI interviewer focusing on awareness, privacy, and security of LLM by 402 participants. Like the static, each participant first completed a self-evaluation before answering any questions.
During the interview, the expert profiler analyzed each response after every question. Based on the current expertise estimate, it selected the next question difficulty (Novice, Basic Knowledge, Advanced Knowledge, or Expert). This internal estimate was not shown to participants. The system uses this expertise profiling to adapt the interview question in real time to improve accuracy. At the end, the profiler produced a final expertise classification for each domain. This final rating was compared with the participant’s self-evaluation to see where they matched or differed. The overall comparison is shown in Table~\ref{tab:profiler_expertise}.





\vspace{-0.2CM}
\begin{table}[h]
\caption{Comparison of Profiler and Human Ratings Across Expertise Levels (L1–L3) for Dynamic Profiling in \%}
\vspace{-0.2CM}
\label{tab:profiler_expertise}
\centering
\scriptsize
\setlength{\tabcolsep}{2pt}
\renewcommand{\arraystretch}{1.1}

\begin{tabular}{
|p{0.19\linewidth}|
 >{\centering\arraybackslash}p{0.01\linewidth}|
 >{\centering\arraybackslash}p{0.085\linewidth}|
 >{\centering\arraybackslash}p{0.085\linewidth}|
 >{\centering\arraybackslash}p{0.085\linewidth}|
 >{\centering\arraybackslash}p{0.085\linewidth}|
 >{\centering\arraybackslash}p{0.085\linewidth}|
 >{\centering\arraybackslash}p{0.085\linewidth}|
}
\hline
\multicolumn{1}{|c|}{\textbf{Case Study}} &
\multicolumn{1}{c|}{\textbf{Same}} &
\multicolumn{3}{c|}{\textbf{Profiler Higher(\%)}} &
\multicolumn{3}{c|}{\textbf{Profiler Lower}}
\\ \cline{3-8}

\multicolumn{1}{|c|}{} &
\multicolumn{1}{c|}{} &
\multicolumn{1}{c|}{\textbf{H1}} &
\multicolumn{1}{c|}{\textbf{H2}} &
\multicolumn{1}{c|}{\textbf{H3}} &
\multicolumn{1}{c|}{\textbf{L1}} &
\multicolumn{1}{c|}{\textbf{L2}} &
\multicolumn{1}{c|}{\textbf{L3}} \\ \hline

Security        & 97 & 1 & 0 & 0 & 2 & 0 & 0 \\ \hline
Privacy         & 95 & 3 & 0 & 0 & 2 & 0 & 0 \\ \hline
LLM Awareness    & 97 & 1 & 0 & 0 & 2 & 0 & 0 \\ \hline

\end{tabular}

\vspace{0.2cm}
{\scriptsize \textit{Note: H1, H2, H3 indicate the Profiler rated participants 1, 2, or 3 levels higher than their self-evaluation. L1, L2, L3 indicate the Profiler rated them 1, 2, or 3 levels lower.}}

\vspace{-0.4CM}
\end{table}

Table~\ref{tab:profiler_expertise} indicates that the profiler’s final rating matches participant self-evaluation in 95–97\% of cases across all domains. Only a very small proportion of participants are rated higher or lower, and most of these differences occur at L1. This suggests that the agentic AI interviewer, combined with the profiler, was effective as it helped converge to the same expert level as the participant’s own belief, especially for those with more experience. The “Profiler Rated Higher” cases indicate under-confident participants, while the “Profiler Rated Lower” cases point to slight overconfidence, mostly among those who see themselves as beginners. Although this pattern cannot be confirmed with full certainty, it is a possible explanation for the observed differences.

To better understand when the profiler converges to the same level as the self-evaluation, we performed a second analysis. For participants whose final profiler rating matched their self-rating, we checked after which question the expertise level first became stable and remained the same until the end of the interview. This reflects the point at which the system has “seen enough” to agree with the participant’s own view. The results are presented in Table~\ref{tab:question_comparison}.

\vspace{-0.2CM}
\begin{table}[h]
\centering
\caption{Identifying Question at Which Profiler Matched the Participant’s Self-Evaluated Expertise Level \& Never Changed in \% }
\vspace{-0.2cm}
\label{tab:question_comparison}
\scriptsize
\setlength{\tabcolsep}{4pt}
\renewcommand{\arraystretch}{0.9}

\begin{tabular}{|c|c|c|c|}
\hline
\textbf{Question Number} & \textbf{Security} & \textbf{Privacy} & \textbf{LLM Awareness} \\ \hline
\textbf{Q1} & 0 & 0 & 0 \\ \hline
\textbf{Q2} & 15 & 3 & 7 \\ \hline
\textbf{Q3} & 86 & 24 & 56 \\ \hline
\textbf{Q4} & 97 & 80 & 87 \\ \hline
\textbf{Q5} & - & 95 & 97 \\ \hline

\end{tabular}
\vspace{-0.4CM}
\end{table}

The results in Table \ref{tab:question_comparison} indicate that the profiler does not find the correct expertise level immediately. It needs a few questions before it can make a stable decision. Almost no participant gets a stable and correct rating after the first question. For Security, most stable matches appear by Questions 3 or 4. This shows that technical topics give clear signals early, making it easier for the profiler to understand the participant’s level. For Privacy and LLM Awareness, the profiler needs more questions. Many participants only reach a stable match by Questions 4 or 5. These topics depend more on interpretation and personal context, so the system requires more evidence before it can judge confidently. This means short assessments are not enough. At least four well-designed questions are needed for accurate results.
To understand how quickly the profiler can form a rough but acceptable judgment of expertise, we also checked when it first reached a rating that was within one level of the participant’s self-evaluation (either H1 above or L1 below). This shows how early the system can approximate a participant’s expertise, even if it has not yet reached the exact final level. Table~\ref{tab:question_comparison_for_level_1} shows the point at which the profiler first achieved this “close enough” match.

\vspace{-0.2CM}
\begin{table}[h]
\centering
\caption{Question at Which first time Profiler Matched +or- level 1 with the Participant’s Self-Evaluated Expertise-Level (H1/L1) in \%}
\vspace{-0.2CM}
\label{tab:question_comparison_for_level_1}
\scriptsize
\setlength{\tabcolsep}{4pt}
\renewcommand{\arraystretch}{0.9}

\begin{tabular}{|c|c|c|c|}
\hline
\textbf{Question Number} & \textbf{Security} & \textbf{Privacy} & \textbf{LLM Awareness} \\ \hline
\textbf{Q1} & 100 & 92 & 97 \\ \hline
\textbf{Q2} & - & 100 & 100 \\ \hline
\textbf{Q3} & - & - & - \\ \hline
\textbf{Q4} & - & - & - \\ \hline
\textbf{Q5} & - & - & - \\ \hline

\end{tabular}
\vspace{-0.4cm}
\end{table}

The results indicate that the profiler can form a near-correct estimate of a participant’s expertise very early, often after the first question. In Security, Privacy, and LLM Awareness, almost all participants reached a “within one level” match immediately. This means the profiler can detect strong early signals from just one response. By the second question, all domains reached 100\% alignment within ±1 level, showing that the system becomes confident quickly. 
We observe mainly that the profiler does not require many questions to reach an initial reasonable estimate, even if more questions are required for an exact match. This indicates that short assessments can be useful for quick screening or early guidance, while longer ones help refine accuracy. We also observed that once the profiler reached the ±1 range, it remained stable and did not widen again.

After examining when the profiler first reached a “close enough” estimate (within ±1 level), we then examined a stricter condition: the exact point at which the profiler first matched the participant’s self-evaluated expertise level. This helps us determine how many questions are required by the system before it attains its final expertise level classification, not just close, but exact.
Table \ref{tab:question_comparison_full} shows the question at which this perfect match first occurred for each domain.

\vspace{-0.2CM}
\begin{table}[h]
\centering
\caption{Identifying the Question at Which Profiler Matched Participant’s Self-Evaluated Expertise Level for the first time in \%}
\vspace{-0.2cm}
\label{tab:question_comparison_full}
\scriptsize
\setlength{\tabcolsep}{4pt}
\renewcommand{\arraystretch}{0.9}

\begin{tabular}{|c|c|c|c|}
\hline
\textbf{Question Number} & \textbf{Security} & \textbf{Privacy} & \textbf{LLM Awareness} \\ \hline
\textbf{Q1} & - & - & - \\ \hline
\textbf{Q2} & 100 & 49 & 79 \\ \hline
\textbf{Q3} & - & 100 & 100 \\ \hline
\textbf{Q4} & - & - & - \\ \hline
\textbf{Q5} & - & - & - \\ \hline

\end{tabular}
\vspace{-0.4CM}
\end{table}

The results in Table \ref{tab:question_comparison_full} indicate that the profiler requires more information before it can match a participant’s exact expertise level. No participant reached a perfect match after the first question. This means one response is not enough for accurate prediction. 
In Security, all participants reached an exact match by Question 2. This indicates that the profiler found technical and structured topics easier and quicker to assess. But in Privacy and LLM Awareness, the system needed two or three questions to reach it; these areas depend more on reasoning, context, and interpretation, so the profiler required more evidence before concluding confidently. We observed mainly that an exact match required slightly more time, especially in subjective or broader domains. While the profiler can make a close guess early, however, it needs a few well-designed questions to reach a match, indicating the importance of having at least 2–3 questions to get reliable results.




\section{Discussion}

The findings indicate that the profiler can quickly detect a participant’s expertise. However, it becomes even more accurate as more responses are provided. This works very well in structured domains like security, where the language is clear and consistent, which explains why security has the highest agreement in both static and dynamic results. However, the profiler found subjective areas such as Privacy and LLM Awareness more difficult because they depended on personal stories, interpretations, and context. As a result, humans and the profiler may understand the same response differently. These disagreements are attributed to the subjective nature of the domain. These domains simply required more evidence before the system could make a confident judgment. Human confidence also affects the results. In Case Study 2, no participant said they were novices, indicating that people find it difficult to admit ignorance. Some rated themselves too high, while others were too careful. These are possible explanations of the differences between the profiler score and self-evaluation rating. However, the profiler uses patterns in the response to evaluate and not self-belief, making its evaluation more objective and consistent. 
The dynamic results exhibited another pattern: the profiler expertise rating became stable earlier. Once it gets close to the participant’s level, it rarely changes. This shows that adaptive questioning works well because it quickly matches the question difficulty with the participant's real-time expertise level, allowing the system to provide a reliable estimate quickly. Generally, the profiler performs best in structured domains, remains consistent even when people misjudge themselves, and benefits greatly from dynamic questioning. This makes it a useful tool to evaluate real-world situations where expertise changes over time and objective evaluation is needed. 

\section{Conclusion}

This study introduced the expert profiler, an explainable AI system that identifies human expertise by processing natural language responses. The results from both static and dynamic tests indicate that the system can classify expertise very accurately across different domains. It performs best when responses are clear and structured, and it stays stable once it attains the correct level. The dynamic interviews showed that the profiler adapts quickly. It needed only a few questions to form a reliable estimate, while also becoming more accurate as more responses are given. This makes it useful for application in real-world situations where a person’s expertise may change over time. However, the profiler still struggles when responses are vague or very subjective. Future improvements will focus on refining the scoring rules, supporting more domains, and optimizing ratings when responses are unclear. Generally, the expert profiler offers a practical and transparent way to understand participant ability levels. This understanding helps create smoother and more meaningful conversations between humans and intelligent systems.

\addtolength{\textheight}{-12cm}   

\end{document}